\documentclass{article}

\usepackage{amssymb}
\usepackage{amsfonts}
\usepackage{graphicx}
\newcommand{\R}{\mathbb{R}}

\begin{document}
\title{Ultrametric Model of Mind, II: Application to 
Text Content Analysis}

\author{Fionn Murtagh \\
Department of Computer Science \\
Royal Holloway University of London \\
Egham, Surrey TW20 0EX, England \\
E-mail fmurtagh@acm.org}

\maketitle

\begin{abstract}
In a companion paper, Murtagh (2012), we discussed how Matte 
Blanco's
work linked the unrepressed unconscious (in the human) to
symmetric logic and thought processes.   We showed how ultrametric topology
provides a most useful representational and computational framework for
this.  Now we look at the extent to which we can find ultrametricity
in text.   We use coherent and meaningful collections of
nearly 1000 texts to show how we can measure inherent ultrametricity.  
On the basis of our findings we hypothesize that inherent ultrametricty 
is a basis for further exploring unconscious thought processes.
\end{abstract}

\section{Introduction}

Any agglomerative hierarchical 
procedure (cf.\ Benz\'ecri, 1979a,b; Lerman, 1981; Murtagh, 1983, 1985) can 
impose hierarchical structure.  Our first aim in this work is to assess 
inherent extent of hierarchical or ultrametric structure.  

We take a large
number of meaningful texts in order to see how they can be distinguished
and/or what other conclusions can be drawn, in regard to their inherent
ultrametricity, or hierarchical structure.  

Our procedure is as follows.  

\begin{enumerate}
\item 
Meaningful component parts of texts are used, such as chapters, reports, 
tales, or very approximately similar sized segments of contiguous 
text.  Our aim is
natural division and also very roughly comparable text component sizes.  
In regard to the latter experimental design choice, very varied text
component lengths are easily accommodated.  

\item  
Then both text units and the word set are projected into a Euclidean
space.  Correspondence analysis allows us to do this.  This projection 
method takes ``profiles'' of counts, or frequencies of occurrence, endowed
with the $\chi^2$ metric, into a Euclidean space.  Both text units 
and words are projected into the same Euclidean space.  All pairwise 
relationships -- between text units, between words, and between both 
sets -- are taken into account in this mapping of the $\chi^2$ 
metric endowed space into the Euclidean metric endowed space.  

\item 
Within each text, based on its Euclidean factor space representation,
we then proceed to investigate how ultrametric it is.  By design,
the ``semantic network'' used and expressed by the Euclidean factor 
space is metric.  How ultrametric it is is the question we raise.  

\item In one study, we look at the words, and seek out ultrametrically-related
words.  

\end{enumerate}

In section \ref{sect2} we discuss how we quantify ultrametricity. 

In section \ref{sect3}, the semantic mapping methodology 
through correspondence analysis  is described.  This is  
the mapping of recorded or input data endowed with 
the $\chi^2$ metric into a Euclidean, factor space.  In this 
Euclidean space, we then pose the question: how ultrametric is the 
given space? 

In section \ref{sect4} we summarize and discuss 
our experimental results.  We characterize texts and collections of 
text, ``fingerprinting'' them in terms of inherent ultrametricity.

In section \ref{sect5} we look within a text, to determine just where 
ultrametricity arises.

\section{Quantifying Ultrametricity}
\label{sect2}

In the companion article (Murtagh, 2012), we described how 
ultrametricity provides a representation (in this sense a model) 
of Matte Blanco's symmetric reasoning.  Symmetric reasoning, as 
we have seen, is associated with repressed or otherwise 
unconscious thought processes.  

Before introducing our method of quantifying ultrametricity, 
we look at some other ways we could do so, albeit in a less
satisfactory way (as we will argue). 

\subsection{Ultrametricity Coefficient of Lerman}

The principle adopted in any constructive assessment of 
ultrametricity is to construct an ultrametric on data and see 
what discrepancy there is between input data and induced 
ultrametric data structure. Quantifying ultrametricity using 
a constructive approach is less than perfect as a solution, 
given the potential complications arising from known problems, 
e.g.\ chaining in single link, and non-uniqueness, or even 
inversions, with other methods. The conclusion here is that 
the ``measurement tool'' used for quantifying ultrametricity 
itself occupies an overly prominent role relative to that 
which we seek to measure. For such reasons, we need an 
independent way to quantify ultrametricity. 

Lerman's (1981) H-classifiability index is as follows.
From the isosceles triangle principle, given a distance $d$ where 
$d(x, y) \neq d(y, z)$ we have $d(x, z) \leq 
\mbox{max} \{ d(x, y), d(y, z) \}$, 
it follows that the largest and second largest of the numbers 
$d(x, y), d(y, z), d(x, z)$ are equal. Lerman's H-classifiability 
measure essentially looks at how close these two numbers (largest, 
second largest) are. So as to avoid influence of distribution of 
the distance values, Lerman's measure is based on ranks (of these 
distances) only.  For further discussion of it, see Murtagh (2004).  

There are two drawbacks with Lerman's index. Firstly, ultrametricity 
is associated with H = 0 but non-ultrametricity is not bounded. In 
extensive experimentation, we found maximum values for H in the 
region of 0.24. The second problem with Lerman's index is that for 
floating point coordinate values, especially in high dimensions, the 
strict equality necessitated for an equilateral triangle is nearly 
impossible to achieve. However our belief is that approximate 
equilateral triangles are very likely to arise in important cases 
of high-dimensional spaces with data points at hypercube vertex 
locations. We would prefer therefore that the quantifying of 
ultrametricity should ``gracefully'' take account of triplets 
which are ``close to'' equilateral. Note that for some authors, 
the equilateral case is considered to be ``trivial'' or a ``trivial 
limit'' (Treves, 1997). For us, however, it is an important case, 
together with the other important case of ultrametricity (i.e., 
isosceles with small base).

\subsection{Ultrametricity Coefficient of Rammal, Toulouse and Virasoro}

The quantifying of how ultrametric a data set is by Rammal et al.\ (1985, 
1986) was influential for us in this work. The Rammal ultrametricity 
index is given by $\sum_{x,y}(d(x, y) - d_c(x, y))/ \sum_{x,y} d(x, y)$ 
where $d$ is the metric distance being assessed, and $d_c$ is the 
subdominant ultrametric. The latter is also the ultrametric associated 
with the single link hierarchical clustering method.  The Rammal et al.\ 
index is bounded by 0 (= ultrametric) and 1. As pointed out in Rammal 
et al.\ (1985, 1986), this index suffers from ``the chaining effect and 
from sensitivity to fluctuations''. The single link hierarchical 
clustering method, yielding the subdominant ultrametric, is, as is 
well known, subject to such difficulties. 

\subsection{Ultrametricity Coefficients of Treves and of Hartman}

Treves (1997) considers triplets of points giving rise to minimal, 
median and maximal distances. In the plot of $d_{\mbox{min}}/d_{\mbox{max}}$ 
against $d_{\mbox{med}}/d_{\mbox{max}}$, the triangular inequality, the 
ultrametric inequality, and the ``trivial limit'' of equilateral triangles, 
occupy definable regions.

Hartmann (1998) considers $d_{\mbox{max}} - d_{\mbox{med}}$. Now, 
Lerman (1981) uses ranks in order to give (translation, scale, etc.) 
invariance to the sensitivity (i.e., instability, lack of robustness) 
of distances. Hartmann instead fixes the remaining distance $d_{\mbox{min}}$.

We seek to avoid, as far as possible, lack of invariance due to use of 
distances. We seek to quantify both isosceles with small base 
configurations, as well as equilateral configurations. Finally, we seek 
a measure of ultrametricity bounded by 0 and 1.

\subsection{Bayesian Network Modeling}

Latent ultrametric distances were estimated by Schweinberger and Snijders
(2003) using a Bayesian and maximum likelihood approach 
in order to represent transitive structures among pairwise 
relationships.  As they state, ``The observed network is generated 
by hierarchically nested latent transitive structures, expressed by 
ultrametrics''.   Multiple, nested transitive structures are at issue.  
``Ultrametric structures imply transitive structures'' and 
as an informal way to characterize ultrametric structures (arising from
embedded clusters, comprising ``friends'' and  ``close friends''):
``Friends are likely to agree, and unlikely to disagree; close 
friends are very likely to agree, and very unlikely to disagree.''

Issues however in the statistical model-based approach to determining
ultrametricity include that convergence to an optimal fit is not 
guaranteed and there can be an appreciable computational requirement.   
Our approach (to be described in the next subsection) in contrast 
is fast and can be achieved through sampling which supposes that 
there is a homogenous ultrametricity pertaining to the data used.  
If sampling is used (for computational reasons) then we assume that 
the text is ``textured'' in the same way throughout, or that it is 
sufficiently ``unified''.  For one theme in regard to content, 
or one origin, or one author, such an assumption seems a reasonable one.     

\subsection{Our Ultrametricity Coefficient}

We define a coefficient of ultrametricity termed $\alpha$ which is 
specified algorithmically as follows.

\begin{enumerate}

\item All triplets of points are considered, with a distance (by 
default, Euclidean) defined on these points. Since for a large number 
of points, $n$, the number of triplets, $n(n - 1)(n - 2)/6$ would be 
computationally prohibitive, we may wish to randomly (uniformly) sample 
coordinates $(i \sim \{ 1..n \}, j \sim \{1..n\}, k \sim \{1..n\})$.

\item We check for possible alignments (implying degenerate triangles) and 
exclude such cases.

\item Next we select the smallest angle as less than or equal to 60 
degrees. (We use the well-known definition of the cosine of the angle 
facing side of length $x$ as: $(y^2 + z^2 - x^2)/2yz.)$ This is our 
first necessary property for being a strictly isosceles ($< 60$ degrees) 
or equilateral ($= 60$ degrees) ultrametric triangle.

\item For the two other angles subtended at the triangle base, we seek 
an angular difference of strictly less than 2 degrees (0.03490656 
radians). This condition is an approximation to the ultrametric 
configuration, based on an arbitrary choice of small angle. This 
condition is targeting a configuration that may not be exactly 
ultrametric but nonetheless is very close to ultrametric.

\item Among all triplets (1) satisfying our exact properties (2, 3) and 
close approximation property (4), we define our ultrametricity coefficient 
as the relative proportion of these triplets. Approximately ultrametric 
data will yield a value of 1. On the other hand, data that is 
non-ultrametric in the sense of not respecting conditions 3 and 4 will 
yield a low value, potentially reaching 0.
\end{enumerate}

In summary, the $\alpha$ index is defined in this way:

Consider a triplet of points, that defines a triangle. If the 
smallest internal angle, $a$, in this triangle is $\leq 60$ degrees, 
and, for the two other internal angles, $b$ and $c$, if $ |b - c| < 2$ 
degrees, then this triangle is an ultrametric one. We look for the 
overall proportion of such ultrametric triangles in our data.

In the Appendix we give the essential pseudo-code used.  

\subsection{What the Ultrametricity Coefficient Reveals}

A wide range of case studies are used in Murtagh (2004) to explore 
this coefficient of ultrametricity.  

It is found that: 
\begin{itemize}
\item the number of 
points (i.e., either words or text components), $n$, does not 
effect the value of the ultrametricity 
coefficient, $\alpha$; 
\item ultrametricity as quantified in this way
increases with sparsity of data encoding (e.g., word presences in 
text components); 
\item ultrametricity 
increases with dimensionality (of either word set, or text component 
set); 
\item dimensionality and spatial (embedding space -- each word 
in the text component space, and each text component in the word
space) 
sparsity, combined, force the tendency towards ultrametricity, but 
the compounding of these two data properties is not as pronounced 
as one might have expected; 
\item and ultrametricity very noticeably
increases with spatial dimensionality.   
\end{itemize}

Furthermore in Murtagh (2004) 
a connection is made with sparse forms of coding in regard to how 
complex stimuli are represented in the cortex.  
Among other implications, this points to the possibility that 
semantic pattern matching is best accomplished through ultrametric 
computation.  

In regard to such ultrametric computation, search can 
benefit from prior ultrametric structuring -- such as through inducing 
a hierarchical clustering on the data -- and then nearest neighbor 
search can be shown to be achievable in constant worst-case 
computational time.  This very powerful result is in keeping with the 
human ability to pattern-match in thought in what appears to be real time.
Murtagh (2004) concludes by noting that it may be the case that human thinking 
is computationally efficient precisely because such computation is 
carried out in an ultrametric space.

So much for the background on the experimental work now to be presented.

With regard to Matte Blanco (1998), the human thinking at issue is
``unrepressed unconscious'' thinking, expressing symmetrical reasoning,
or more the symmetrical mode of being.   This is one facet of the 
bi-logical system in the human mind process.  

\section{Semantic Mapping: Mapping Interrelationships into a Euclidean,
Factor Space}
\label{sect3}

We employ correspondence analysis for metric embedding,
followed by determination of the extent of  ultrametricity, in factor
space, based on the $\alpha$ coefficient of ultrametricity.  Our motivation 
for using precisely this Euclidean embedding is as follows.  Our input 
data is in the form of frequencies of occurrence.  Now, a Euclidean distance
defined on vectors with such values is not appropriate.  

The $\chi^2$ distance
is an appropriate weighted Euclidean distance for use with such data
(Benz\'ecri, 1979; Murtagh, 2005b).  
Consider texts $i$ and $i'$ crossed by words $j$.  Let $k_{ij}$ be the number of
occurrences of word $j$ in text $i$.  Then, omitting a constant, 
the $\chi^2$ distance between texts $i$ and $i'$ is given by 
$ \sum_j 1/k_j ( k_{ij}/k_i - k_{i'j}/k_{i'} )^2$.  The weighting term is 
$1/k_j$.  The weighted Euclidean distance is between the {\em profile} 
of text $i$, viz.\ $k_{ij}/k_i$ for all $j$, and the analogous 
{\em profile} of text $i'$.

%

\subsection{Correspondence Analysis: 
Mapping $\chi^2$ into Euclidean Distances}

As a dimensionality reduction technique 
correspondence analysis is particularly appropriate for handling 
frequency data.  As an example of the latter, frequencies of word
occurrence in text will be studied below.  

The given contingency table (or numbers of occurrence) 
data is denoted $k_{IJ} =
\{ k_{IJ}(i,j) = k(i, j) ; i \in I, j \in J \}$.  $I$ is the set of text
indexes, and $J$ is the set of word indexes.  We have
$k(i) = \sum_{j \in J} k(i, j)$.  Analogously $k(j)$ is defined,
and $k = \sum_{i \in I, j \in J} k(i,j)$.  Next, $f_{IJ} = \{ f_{ij}
= k(i,j)/k ; i \in I, j \in J\} \subset \R_{I \times J}$,
similarly $f_I$ is defined as  $\{f_i = k(i)/k ; i \in I, j \in J\}
\subset \R_I$, and $f_J$ analogously.  What we have described here is 
taking numbers of occurrences into relative frequencies.

The conditional distribution of $f_J$ knowing $i \in I$, also termed
the $j$th profile with coordinates indexed by the elements of $I$, is:

$$ f^i_J = \{ f^i_j = f_{ij}/f_i = (k_{ij}/k)/(k_i/k) ; f_i \neq 0 ;
j \in J \}$$ and likewise for $f^j_I$.  

Note that the input data values here are always non-negative reals.  The 
output factor projections (and contributions to the principal directions 
of inertia) will be reals.  

\subsection{Input: Cloud of Points Endowed with the Chi Squared Metric}

The cloud of points consists of the couple: profile coordinate and mass.
We have $ N_J(I) = \{ ( f^i_J, f_i ) ; i  \in I \} \subset \R_J $, and
again similarly for $N_I(J)$.

The moment of inertia is as follows: 
$$M^2(N_J(I)) = M^2(N_I(J)) = \| f_{IJ} - f_I f_J \|^2_{f_I f_J} $$
\begin{equation}
= \sum_{i \in I, j \in J} (f_{ij} - f_i f_j)^2 / f_i f_j
\end{equation}
The term  $\| f_{IJ} - f_I f_J \|^2_{f_I f_J}$ is the $\chi^2$ metric
between the probability distribution $f_{IJ}$ and the product of marginal
distributions $f_I f_J$, with as center of the metric the product
$f_I f_J$.  Decomposing the moment of inertia of the cloud $N_J(I)$ -- or 
of $N_I(J)$ since both analyses are inherently related -- furnishes the 
principal axes of inertia, defined from a singular value decomposition.

\subsection{Output: Cloud of Points Endowed with the Euclidean 
Metric in Factor Space}

From the initial frequencies data matrix, a set of probability data,
$f_{ij}$, is defined by dividing each value by the grand total of all
elements in
the matrix.  In correspondence analysis,
each row (or column) point is considered to have an
associated weight.  The weight of the $i$th row point is given
by $f_i = \sum_j f_{ij}$, and the weight of the $j$th column point
is given by $f_j = \sum_i f_{ij}$. We consider the row points to have
coordinates ${f_{ij} / f_i}$, thus allowing points of the same
{\em profile} to be identical (i.e., superimposed). The following weighted
Euclidean distance, the $\chi^2$ distance, is then used between row
points:
$$ d^2(i,k) = \sum_j {1 \over f_j} \left( {f_{ij} \over f_i} -
                                     {f_{kj} \over f_k} \right)^2 $$
and an analogous distance is used between column points.

The mean row point is given by the weighted average of all row
points:
$$ \sum_i f_i {f_{ij} \over f_i} = f_j$$
for $j = 1, 2, \dots, m$.  Similarly the mean column profile has
$i$th coordinate $f_i$.

We
first consider the projections of the $n$
profiles in $\R^m$ onto an axis, ${\bf u}$.  This is given by
$$ \sum_j {f_{ij} \over f_i} {1 \over f_j} u_j$$ for all $i$ (note
the use of the scalar product here).  For details on determining the 
new axis, ${\bf u}$, see Murtagh (2005).

The  projections of points onto
axis ${\bf u}$ were with respect to the ${1 / f_i}$ weighted Euclidean
metric.  This makes interpreting projections very difficult from a
human/visual point of view, and so it is more natural to present results
in such a way that projections can be simply appreciated.  Therefore
{\em factors} are defined, such that the projections of row vectors
onto factor ${\bf \phi}$ associated with axis ${\bf u}$ are given by
$$\sum_j {f_{ij} \over f_i} \phi_j$$ for all $i$.  Taking $$\phi_j =
{1 \over f_j} u_j$$ ensures this and projections onto ${\bf \phi}$
are with respect to the ordinary (unweighted) Euclidean distance.

An analogous set of relationships hold in $\R^n$ where the best
fitting axis, ${\bf v}$, is searched for.  A simple mathematical
relationship holds between ${\bf u}$ and ${\bf v}$, and between
${\bf \phi}$ and ${\bf \psi}$ (the latter being the factor associated
with axis or eigenvector ${\bf v}$):
$$ \sqrt{\lambda} \psi_i = \sum_j {f_{ij} \over f_i} \phi_j $$
$$ \sqrt{\lambda} \phi_j = \sum_i {f_{ij} \over f_j} \psi_i $$
These are termed {\em transition formulas}. 
 Axes ${\bf u}$
and ${\bf v}$, and factors ${\bf \phi}$ and ${\bf \psi}$, are
associated with eigenvalue $\lambda$ and best fitting higher-dimensional
subspaces are associated with decreasing values of $\lambda$ (see Murtagh,
2005b, for further details).

In this work, $\phi_j$ are coordinates of words in the new, factor 
and Euclidean, space.  The $\psi_i$ are coordinates of text segments 
in the factor space.  
In the Euclidean, factor space, the transition formulas have the 
following interpretation.  Each text point is the weighted average of
its associated word points.  Similarly, each word is located at the 
center of gravity of its associated texts.   In this way the factor 
space of the text segments and the factor space of the words furnish 
one semantic space.  

\subsection{Conclusions on Correspondence Analysis and Introduction to the 
Numerical Experiments to Follow}
\label{sect34}

Some important points for the analyses to follow are -- firstly in relation 
to correspondence analysis: 

\begin{enumerate}

\item From numbers of occurrence data we always get (by design) 
a Euclidean embedding
using correspondence analysis.  The factors are embedded in a Euclidean 
metric.  

\item Due to centering the data, the 
numbers of factors, i.e.\ number of non-zero eigenvalues, are
given by one less than the minimum of the number of observations studied
(indexed by set $I$) and the number of variables or attributes used 
(indexed by set $J$).  

\item The number of dimensions in factor space may be less than full rank
if there are linear dependencies present.

\item In the experiments to follow in the next section, we have $n < m$ 
always, implying that inherent (full rank) 
dimensionality of the projected Euclidean 
factor space is $n - 1$.  

\item We also take $m = 1000, 2000$ and the full attribute set (say, 
$m_{\rm tot}$) in each case, where the attributes are ordered in terms of 
decreasing marginal frequency.  In other words, we take the 1000 most
frequent words to characterize our texts; then the 2000 most frequent words; 
and finally all words.  Since $n < m$ it is not surprising that 
similar results are found irrespective of the value of $m$.  
The inherent, projected, Euclidean, factor space dimensionality is the 
same in each case, viz., $n - 1$.  

\item From the previous remark, viz.\ that the results obtained for the
$m = 1000, 2000$, and all most frequent  words, are of the same inherent 
dimensionality we motivate our use of these different characterizations of
the text set by the need to study the stability of 
our results.  We will show quite convincingly that our results are 
characteristic of the texts used, in each case, and are not ``one off''
or arbitrary.  


\end{enumerate}

Some important points related to our numerical assessments below, in 
relation to data used, determining of ultrametricity coefficient, 
and software used, are as follows.

\begin{enumerate}

\item 
In line with one tradition of textual analysis associated with Benz\'ecri's
correspondence analysis (see Chapter 5, ``Content analysis of text'', in 
Murtagh, 2005b) we take the unique full words and
rank them in order of importance.  Thus for the Brothers Grimm work,
below, we find function words: ``the'', 19,696 occurrences; ``and'',
14,582 occurrences; ``to'', 7380 occurrences; ``he'', 5951 occurrences; 
``was'', 4122 occurrences; and so on.  Last three, with one occurrence each:
``yolk'', ``zeal'', ``zest''.   

\item The $\alpha$ ultrametricity coefficient is based on triangles. Now, 
with $n$ graph nodes we have $O(n^3)$ possible triangles which is 
computationally prohibitive, so we instead sample.  The means and 
standard deviations below are based on 2000 random triangle vertex
realizations, repeated 20 times; hence, in each case, in total 40,000 
random selections of triangles.  

\item All text collections reported on below (section \ref{sect4}) 
are publicly accessible (and web addresses are cited).  All texts were
obtained by us in straight (ascii) text format.
   
The preparation of the input data was carried out with programs
written in C, 
and available at www.correspondances.info (accompanying 
Murtagh, 2005b).  The correspondence analysis software was written in
the public  R statistical software environment  
(www.r-project.org, again see Murtagh, 2005b) and is available at this same 
web address.  Some 
simple statistical calculations were carried out by us also 
in the R environment.  

\end{enumerate}

\section{Determining Ultrametricity through 
Interrelationships between Text Units based on Shared Words}
\label{sect4}

We use in all over 900 short texts, given by short stories, or chapters,
or short reports.  All are in English.  Unique words are determined 
through delimitation by white space and by punctuation characters
with no distinction of upper and lower case.  In
all, over one million words are used in our studies of these texts.  

We carried out some assessments of Porter stemming (Porter, 1980)
as an alternative 
to use of whitespace- or punctuation-delimited words, without much 
difference in our findings.   

\subsection{Brothers Grimm}

As a homogeneous collection of texts we take 209 fairy tales of the Brothers 
Grimm (Ockerbloom, 2003), 
containing 7443 unique (in total 280,629) space- or 
punctuation-delimited words.  Story lengths were between 650 and 44,400 words.

To define a semantic context of increasing
resolution we took the most frequent 1000 words, followed by the most frequent
2000 words, and finally all 7443 words.  
We constructed a cross-tabulation of numbers of occurrences of
each word in each one of the 209 fairy tales.  This led therefore to a 
set of frequency tables (contingency tables) 
of dimensions: $209 \times 1000,
209 \times 2000$ and $209 \times 7443$.    
The factor space, of dimension $209 - 1 = 208$ (cf.\ subsection
\ref{sect34}), is Euclidean, so 
the correspondence analysis can be said to be
a mapping from the $\chi^2$ metric into a Euclidean metric space.


\begin{table}
\caption{Coefficient of ultrametricity, $\alpha$.
Input data: frequencies of occurrence matrices defined on the 209 texts 
crossed by: 
1000, 2000, and all = 7443, words.  
$\alpha$ (ultrametricity coefficient) based
on factors: i.e., factor projections resulting 
from correspondence analysis, with Euclidean distance used between each 
pair of texts in factor space, of dimensionality 208.  
}
\label{tabcorrb}
\begin{center}
\setlength{\tabcolsep}{1mm}
\begin{tabular}{|crrrr|} \hline 
      &  \multicolumn{3}{c}{209 Brothers Grimm fairy tales}  &  \\ \hline
Texts & Orig.Dim. & FactorDim. & $\alpha$, mean & $\alpha$, sdev. \\ \hline 
209   &  1000     & 208  &  0.1236   &  0.0054 \\
209   &  2000     & 208  &  0.1123   &  0.0065 \\
209   &  7443    & 208  &  0.1147   &  0.0066 \\ \hline
\end{tabular}
\end{center}
\end{table} 




Table \ref{tabcorrb} (columns 4, 5) 
shows remarkable stability of the $\alpha$ ultrametricity
coefficient results, and such stability will be seen in all further results 
to be presented below.  In the table, means and standard deviations 
were calculated in each case from 2000 random triangles, repeated 20 times
(cf.\ subsection \ref{sect34}).  
The ultrametricity is not high for the Grimm 
Brothers' data: we recall that an $\alpha$ value of 0 means no triangle is 
isosceles/equilateral.  We see that there is very little ultrametric
(hence hierarchical) structure in the Brothers Grimm data (based on our 
particular definition of ultrametricity/hierarchy).

\subsection{Jane Austen}

To further study stories of a general sort, we use some works of the 
English novelist, Jane Austen.  

\begin{enumerate}
\item {\em Sense and Sensibility} (Austen, 1811), 
50 chapters = files, chapter lengths from 1028 to 5632 words.
\item {\em Pride and Prejudice} (Austen, 1813), 
61 chapters each containing between 683 and 5227 words. 
\item {\em Persuasion} (Austen, 1817), 24 chapters,
chapter lengths 1579 to 7007 words.
\item {\em Sense and Sensibility} split into 131 separate 
texts, each containing around 1000 words
(i.e., each chapter was split into files containing 5000 or fewer characters).
We did this to check on any influence by the size (total number of words) of
the text unit used (and we found no such influence).  
\end{enumerate}  

In all there were 266 texts containing a total of 9723 unique words.  We 
looked at the 1000, 2000 most frequent, and all 9723, words to 
characterize the texts by frequency of occurrence.


\begin{table}
\caption{Coefficient of ultrametricity, $\alpha$.  
Input data: frequencies of occurrence matrices defined on the 266 texts 
crossed by: 
1000, 2000, and all = 9723, words.  
$\alpha$ (ultrametricity coefficient) based
on factors: i.e., factor projections resulting 
from correspondence analysis, with Euclidean distance used between each 
pair of texts in factor space.  
Dimensionality of latter is necessarily $ \leq 266 -1$,
adjusted for 0 eigenvalues = linear dependence. 
}
\label{tabcorr2b}
\begin{center}
\setlength{\tabcolsep}{1mm}
\begin{tabular}{|crrrr|} \hline 
  & \multicolumn{3}{c}{266 Austen chapters or partial chapters} & \\ \hline
Texts & Orig.Dim. & FactorDim. & $\alpha$, mean & $\alpha$, sdev. \\ \hline 
266   &  1000     & 261  &  0.1455   &  0.0084 \\
266   &  2000     & 262  &  0.1489   &  0.0083 \\
266   &  9723     & 263  &  0.1404   &  0.0075 \\ \hline
\end{tabular}
\end{center}
\end{table} 

Table \ref{tabcorr2b}, again displaying very stable $\alpha$ values, 
indicates
that the Austen corpus is a small amount more ultrametric than the Grimms'
corpus, Table \ref{tabcorrb}.

\subsection{Air Accident Reports}

We used air accident reports to explore documents with very particular,
technical, vocabulary.  
The NTSB aviation accident database  
(Aviation Accident Database and Synopses, 2003)
contains information 
about civil aviation accidents in the United States and elsewhere.
We selected 50 reports.  Examples of two such reports used
by us: occurred Sunday, January 02, 2000 in Corning, AR,
aircraft Piper PA-46-310P, injuries -- 5 uninjured; occurred Sunday,
January 02, 2000 in Telluride, TN, aircraft: Bellanca BL-17-30A,
injuries -- 1 fatal.  In the 50 reports, there were 55,165 words.
Report lengths ranged between approximately 2300 and 28,000 words. The
number of unique words was 4261.

Example of the start of our 30th report: {\em On January 16, 2000, about
1630 eastern standard time (all times are eastern standard time,
based on the 24 hour clock), a Beech P-35, N9740Y, registered to a
private owner, and operated as a Title 14 CFR Part 91 personal
flight, crashed into Clinch Mountain, about 6 miles north of
Rogersville, Tennessee. Instrument meteorological conditions prevailed
in the area, and no flight plan was filed. The aircraft incurred
substantial damage, and the private-rated pilot, the sole occupant,
received fatal injuries. The flight originated from Louisville,
Kentucky, the same day about 1532.}


\begin{table}
\caption{Coefficient of ultrametricity, $\alpha$.  
Input data: frequencies of occurrence matrices defined on the 50 texts 
crossed by: 
1000, 2000, and all = 4261, words.  
$\alpha$ (ultrametricity coefficient) based
on factors: i.e., factor projections resulting 
from correspondence analysis, with Euclidean distance used between each 
pair of texts in factor space.  
Dimensionality of latter is necessarily $ \leq 50 -1$,
with an additional adjustment made for one 0-valued eigenvalue,
implying linear dependence. 
}
\label{tabcorr4b}
\begin{center}
\setlength{\tabcolsep}{1mm}
\begin{tabular}{|crrrr|} \hline 
  & \multicolumn{3}{c}{50 aviation accident reports} &  \\ \hline
Texts & Orig.Dim. & FactorDim. & $\alpha$, mean & $\alpha$, sdev. \\ \hline 
50   &  1000     & 48  &  0.1338   &  0.0077 \\
50   &  2000     & 48  &  0.1186   &  0.0058 \\
50   &  4261     & 48  &  0.1154   &  0.0050 \\ \hline
\end{tabular}
\end{center}
\end{table}

In Table \ref{tabcorr4b} we find ultrametricity values that are marginally
greater than those found for the Brothers Grimm (Table \ref{tabcorrb}).  It 
could be argued that the latter, too, uses its own technical 
vocabulary.   We would need to use more data to see if we can clearly 
distinguish between the (small) ultrametricity levels of these two 
corpora.

\subsection{DreamBank}
\label{sect44}

With dream reports (i.e., reports by individuals on their remembered 
dreams) we depart from a technical vocabulary, and instead raise the 
question as to whether dream reports can perhaps be considered as types
of fairy tale or story, or even akin to accident reports.  

From the Dreambank repository (Domhoff, 2003; DreamBank, 2004; Schneider
and Domhoff, 2004) 
we selected the following collections:

\medskip

(1) ``Alta: a detailed dreamer,'' in period 1985--1997, 422 dream reports.

(2) ``Chuck: a physical scientist,''  in period
1991--1993,  75 dream reports.

(3) ``College women,'' in period 1946--1950,  681 dream reports.

(4) ``Miami Home/Lab,''  in period  1963--1965,  445 dream reports.

(5) ``The Natural Scientist,''  1939,  234 dream reports.

(6) ``UCSC women,''  1996,  81 dream reports.

\medskip

To have adequate length reports, we requested report sizes of between
500 and 1500 words.  With this criterion, from (1) we obtained 118 reports,
from (2) and (6) we obtained no reports, from (3) we obtained 15 reports,
from (4) we obtained 73 reports, and finally from (5) we obtained 8 reports.
In all, we used 214 dream reports, comprising 13696 words.


As an example, here is the start of the 100th (for us) 
report: {\em I'm delivering a car to a man --
something he's just bought, a Lincoln
Town Car, very nice. I park it and go down the street to find him -- he
turns out to be an old guy, he's buying the car for nostalgia -- it turns
out to be an old one, too, but very nicely restored, in excellent
condition. I think he's black, tall, friendly, maybe wearing overalls. I
show him the car and he drives off. I'm with another girl who drove
another car and we start back for it but I look into a shop first -- it's
got outdoor gear in it -- we're on a sort of mall, outdoors but the shops
face on a courtyard of bricks. I've got something from the shop just
outside the doors, a quilt or something, like I'm trying it on, when
it's time to go on for sure so I leave it on the bench. We go further,
there's a group now, and we're looking at this office facade for the
Honda headquarters.}

With the above we took another set of dream reports, from one individual,
Barbara Sanders.  A more reliable (according to DreamBank, 2004) set of
reports comprised 139 reports, and a second comprised 32 reports.  In all
171 reports were used from this person.  Typical lengths were about 2500
up to 5322.  The total number of words in the Barbara Sanders set of
dream reports was 107,791.


\begin{table}
\caption{Coefficient of ultrametricity, $\alpha$.  
Input data: frequencies of occurrence matrices defined on the 384 texts 
crossed by: 
1000, 2000, and all = 11441, words.  
$\alpha$ (ultrametricity coefficient) based
on factors: i.e., factor projections resulting 
from correspondence analysis, with Euclidean distance used between each 
pair of texts in factor space, of dimensionality $ 385 -1 = 384$.  
}
\label{tabcorr3b}
\begin{center}
\setlength{\tabcolsep}{1mm}
\begin{tabular}{|crrrr|} \hline 
 & \multicolumn{3}{c}{385 dream reports}  & \\ \hline
Texts & Orig.Dim. & FactorDim. & $\alpha$, mean & $\alpha$, sdev. \\ \hline 
385   &  1000     & 384  &  0.1998   &  0.0088 \\
385   &  2000     & 384  &  0.1876   &  0.0095 \\
385   &  11441    & 384  &  0.1933   &  0.0087 \\ \hline
\end{tabular}
\end{center}
\end{table} 

First we analyzed all dream reports, furnishing Table \ref{tabcorr3b}. 

In order to look at a more homogeneous subset of dream reports, we 
then analyzed separately 
the Barbara Sanders set of 171 reports, leading to Table \ref{tabcorr333b}.  
(Note that this analysis is on a subset of 
the previously analyzed dream reports, Table \ref{tabcorr3b}).  
The Barbara Sanders subset of 171 reports contained 7044
unique words in all.

Compared to Table \ref{tabcorr3b} based on the entire dream report 
collection, Table \ref{tabcorr333b} which is based on one person 
shows, on average, higher ultrametricity levels.  It is interesting to note
that the dream reports, collectively, are higher in ultrametricity level 
than our previous values for $\alpha$; and that the ultrametricity level is 
raised again when the data used relates to one person.  

We carried out a preliminary study of James Joyce's {\em Ulysses}, comprising 
304,414 words in total.  We broke this text into 183 separate sequential
files, 
comprising approximately between 1400 and 2000 words each.  The number of 
unique words in these 183 files was found to be 28,649 words.  The 
ultrametricity $\alpha$ values for this collection of 183 Joycean texts 
were found to be less than the Barbara Sanders values, but higher than the 
global set of all dream reports.  


\begin{table}
\caption{Coefficient of ultrametricity, $\alpha$.  
Input data: frequencies of occurrence matrices defined on the 171 texts 
crossed by: 
1000, 2000, and all = 7044, words.  
$\alpha$ (ultrametricity coefficient) based
on factors: i.e., factor projections resulting 
from correspondence analysis, with Euclidean distance used between each 
pair of texts in factor space, of dimensionality $ 171 -1 = 170$. 
}
\label{tabcorr333b}
\begin{center}
\setlength{\tabcolsep}{1mm}
\begin{tabular}{|crrrr|} \hline 
 & \multicolumn{3}{c}{171 Barbara Sanders dream reports}  & \\ \hline
Texts & Orig.Dim. & FactorDim. & $\alpha$, mean & $\alpha$, sdev. \\ \hline 
171   &  1000     & 170  &  0.2250   &  0.0089 \\
171   &  2000     & 170  &  0.2256   &  0.0112 \\
171   &  7044     & 170  &  0.2603   &  0.0108 \\ \hline
\end{tabular}
\end{center}
\end{table}

\section{Ultrametric Properties of Words}
\label{sect5}

\subsection{Objectives and Choice of Data}

The foregoing analyses have been based on text segments and their 
interrelationships.   As noted earlier however, correspondence analysis
projects both text segments and words, both endowed initially with 
the $\chi^2$ metric, into the one Euclidean space. As also observed, this
Euclidean factor space takes all interrelationships into consideration.
We stress too that we are {\em not} using a reduced dimensionality 
approximation of the factor space, as is often done so as to filter
out from 
the data what is considered to be noise.  Instead we use the full 
Euclidean and factor space dimensionality because we wish to study 
the data as given to us but simply endowed
with the usual (i.e.\ unweighted) Euclidean distance.  
(We also assume no recoding of the input data such as through
complete disjunctive or fuzzy or other forms of coding which could
turn the $\chi^2$ distance right away into a Euclidean distance: see Murtagh, 
2005, for discussion of such input data recoding.)

In order to have a text that ought to contain vestiges of ultrametricity 
because of subconscious thinking, admittedly subconscious thinking that was 
afterwards reported on in a fully conscious way, we took the 
Barbara Sanders dream reports.  In section \ref{sect44} we have seen how 
ultrametric we found this data to be.  In discussion of this data 
provided in Domhoff (2002) he notes that there is ``astonishing 
consistency'' shown in dreams such as these over long periods of time.

%
%

Taking a set of 139 of the Barbara Sanders dream reports, as used 
in section \ref{sect44}, we used the 2000 most frequently occurring
words used in these dream reports including function words.   Then we 
took 30 words to carry out some experimentation with their ultrametric
properties.  These are listed in Table \ref{tabultraBS}.  We selected 
these words to have some personal names, some words that could be 
metaphors for the commonplace or the fearful, and some words that could be
commonplace and hence banal.  



Two sets of experiments were carried out.  For both 
experiments, the 30 selected words were
given by their Euclidean space vectors resulting from the correspondence
analysis, carried out on the 139 dream reports $\times$ 2000 words.  So 
the 30 selected words are vectors in a space of dimensionality min(139 $-$ 1,
2000 $-$ 1) = 138.  In the first experiment the ultrametric triangles formed
between triples solely on the 30-word set were determined.  So for each word,
the number of triangles checked was $1 \times (30 - 1) \times (30 - 2) / 2 =
406$.   In the second experiment, the ultrametric triangles formed between the 
selected word and all pairs of the full set of 2000 words were used.  The 
number of triangles checked for each word was $ 1 \times (2000 - 1) \times 
(2000 - 2) / 2 = 1997001$.   However some of these have overlapping 
points, implying zero distances. Rather than 1997001 triangles to be 
checked for each of the 2000 words, instead 1996997 involved no zero-valued
distance.  

\begin{table}
\begin{tabular}{|crrr|} \hline
Selected  &  \# UM cases  & \# UM cases &  Previous col.: \\
words     &  300-word set &  2000-word set & H(igh), L(ow) \\ 
          &  (total triangles:  &  (total triangles: & (defined by \\
          &   406)              &   1,996,997)       & median)   \\ \hline
Tyler    &   24 & 132193 & L \\
Jared    &   19 & 126617 & L \\
car      &   14 &  99631 & L \\
road     &   14 & 107924 & L \\
Derek    &   15 & 187027 & H \\
John     &   17 & 137802 & H \\
Jamie    &   24 & 130304 & L \\
Peter    &   48 & 134052 & L \\
arrow    &   21 & 133917 & L \\
dragon   &   24 & 170157 & H \\
football &   18 & 127036 & L \\
Lance    &   22 & 166112 & H \\
room     &    5 &  65332 & L \\
bedroom  &   13 & 129206 & L \\
family   &   26 & 165286 & H \\
game     &   19 & 171561 & H \\
Mabel    &   60 & 135192 & L \\
crew     &   31 & 128655 & L \\
director &   19 & 143889 & H \\
assistant &  58 & 135250 & L \\
balloon   &  23 & 138154 & H \\
ship      &  18 & 154960 & H \\
balloons  &  23 & 147757 & H \\
pudgy     &  41 & 131698 & L \\
Valerie   &  17 & 161231 & H \\
dolly     &  20 & 140355 & H \\
cat       &  11 & 144958 & H \\
gun       &  20 & 166147 & H \\
Howard    &  28 & 172760 & H \\
horse     &  52 & 132675 & L \\ \hline
\end{tabular}
\caption{Results found on the Barbara Sanders set of 139 dream reports 
for 30 selected words.  ``\# UM cases'' $=$ number of ultrametric 
(triangle) cases.
The numbers of ultrametric-respecting triangles 
were sought.  Such triangles are either equilateral or isosceles with small
base using Euclidean input data.}
\label{tabultraBS}
\end{table}

\subsection{General Discussion of Ultrametricity of Words}

General discussion of Table \ref{tabultraBS} follows.  

\begin{itemize}
\item Note the semantic similarity between ``road'' and ``car'', clearest when dealing
with the 30-word set in isolation, rather than 
the 30-word set in the full 2000-word
context.

\item
Similarly note the semantic similarity between ``balloon'' and ``balloons''. 

\item
Regarding the following words, our information is from Domhoff (2002; and 
further discussion is in Domhoff, 2012).  

\item
``Derek'' (``H'', high number of ultrametric relationships found with this word): 
the dreamer, Barbara Sanders, had a former relationship with him.

\item
``Mabel'' 
(``L'', relatively low number of ultrametric relationships):
co-worker.  The relatively low number of ultrametric relationships found was based 
on the full 2000-word set, -- cf.\ 135192 cases; but when the restricted 30-word 
set alone was used in isolation a much larger relative number of 60 ultrametric cases 
was noted.

\item
``cat'' (``H'', high number of ultrametric relationships):
Barbara Sanders 
has several cats, treats them well in real life, thinks of them as mistreated in 
dreams.

\item
``gun'' (``H'', high number of ultrametric relationships):
Her dreams seem to infer that she used guns when young, but this was 
not in fact the case. 

\item
``Howard'' (``H'', high number of ultrametric relationships): ex-husband.

\item
``horse'' (``L'', relatively low number of ultrametric relationships):
she rides in dreams, fears in real life.
\end{itemize}

\subsection{Conclusions on the Word Analysis}

Derek, with whom there was a former relationship, and Howard, an ex-husband of 
Barbara Sanders, both figure relatively highly in terms of ultrametric relationships,
as can be seen in Table \ref{tabultraBS}.   Admittedly these 
ultrametric-respecting triplets are few in number compared to the total number
of these triplets, viz.\ 1,996,997 or nearly two million per word.  

The distribution of the ultrametric-respecting triangles in a data set such as 
this allows us to assess the statistical significance 
of ultrametricity of any given word.  Our approach is to determine 
the empirical distribution function (rather than, say, a stochastic 
graph model).  Justification is to have a data-driven baselining rather
than an a priori model for the data.  
Therefore we 
looked at the approximately two million triangles that are with reference to 
any word among the 2000 words retained.  

Hence for this distribution we 
used approximately 4000 million triangles.
With reference to the third column, therefore, of Table \ref{tabultraBS}, the 
very maximum number of ultrametric-respecting triangles with account taken of 
all 2000 words was found as 206,496.   To determine this we checked all
2000 words.  The very minimum number of ultrametric-respecting
triangles is 31,346.  These correspond respectively to our $\alpha$ ultrametricity
coefficients of 0.103403 and 0.015697.   

Note that the results of 
Table \ref{tabcorr333b} were based on the dream reports.  While the word 
results are different, this just points to different ultrametricity properties
in the two dual spaces.  Our provisional conclusion is regard to the 
difference in ultrametricity properties in the dual spaces is that it 
may be useful to experiment with content tagging (see the Hall/Van de 
Castle coding system, described at Dreambank, 2004).   

The measured ultrametricity of the word ``Derek'' (former 
relationship) is at the 73.887 percentile,
implying a 26\% chance of being bettered in this data.  The measured 
ultrametricity of the word ``Howard'' (ex-husband) 
is at the 65.583 percentile.   

Our objective in this word analysis has been to indicate the type of vantage
points that can be opened up through the topology analysis that has been our
focus in this work.

\section{Conclusion}

We studied a range of text corpora, comprising about 1000 texts, or 
text segments, containing over 1.3 million words.  We found very 
stable ultrametricity 
quantifications of the text collections, across numbers of most frequent 
words used to characterize the texts, and sampling of triplets of texts.  
Notable aspects of our data analysis include: full inherent dimensionality 
used; full set of words used too in many cases; and finally in 
section \ref{sect5}, sampling was not used but rather exhaustive processing.

We found that in all cases (save, perhaps, the Brothers Grimm versus 
air accident reports) there was a clear distinction between the ultrametricity
values of the text collections.  

Some very intriguing ultrametricity characterizations were found in our
work.  For example, we found that the technical vocabulary of air accidents 
did not differ greatly in terms of inherent ultrametricity compared to the 
Brothers Grimm fairy tales.  Secondly we found that novelist Austen's 
works were clearly 
distinguishable from the Grimm fairy tales.  Thirdly we found 
dream reports to have higher ultrametricity level than the other 
text collections.  

Values of our $\alpha$ ultrametricity coefficient were small but revealing
and valuable, in the sense of being consistent (i.e.\ with small variability)
and being discriminatory (i.e.\ between genres). 

It is interesting to speculate on how one would 
exploit the ``strands'' or ``threads'' of ultrametricity, and hence 
hierarchical structure, that we find.  We use these metaphors (``strands'',
``threads'') with care 
because an ultrametric triangle possibly shares vertices with a non-ultrametric
triangle.   

All in all however we have presented excellent proof of concept that from 
empirical -- textual -- data we can determine measures of ultrametricity,
or hierarchical symmetry.  To that extent we have developed an operational
procedure for ranking (at least as a good first stage of processing) 
manifestations of reasoning in terms of Matte Blanco's 
symmetric, on the one hand, and asymmetric, on the other hand, logic.  

\section*{Appendix: Pseudo-Code for Assessing Ultrametric-Respecting 
Triplet}

Assumed: vectors $i, j, k$ are in a Euclidean space. 

\begin{itemize}

\item 
For all triplets $i, j, k$, consider their Euclidean distances,
$d_1 = d(i,j); d_2 = d(j,k); d_3 = d(i,k)$.  

\item Set $\epsilon$ = $1.0e^{-10}$

\item 
Exclude near-0 distances: 
only 
if ($d_1 > \epsilon$ \& $d_2 > \epsilon$ \& $d_3 > \epsilon$) do 
the following:

\item
Determine cosines of the three angles in the triangle using scalar product,
denoted $\cdot$.

$c_1  = (d_1 \cdot d_1 + d_2 \cdot d_2 - d_3 \cdot d_3) / 
   (2.0 \cdot d_1 \cdot d_2) $

$c_2  = (d_2 \cdot d_2 + d_3 \cdot d_3 - d_1 \cdot d_1) / 
   (2.0 \cdot d_2 \cdot d_3) $

$c_3  = (d_1 \cdot d_1 + d_3 \cdot d_3 - d_2 \cdot d_2) / 
   (2.0 \cdot d_1 \cdot d_3) $

Order these and we will take the case such that $c_1 \leq c_2 \leq c_3 $

\item 
Wanting the largest cosine to correspond to an angle 
less than 60 degrees and greater than 0 degree,
implying that we have a sufficient condition for an isosceles with small base 
triangle, we require the following.  Allowing less than or equal to 
60 degrees encompasses the equilateral triangle case. 
Angle and cosine vary inversely.   

\item 
if ($c_3 \geq 0.5$ \& $c_3 < 1.0$) then:

Assess difference of angles.  Note:  2 degrees = 0.03490656 radians.

$a_1$ = arccos($c_1$)

$a_2$ = arccos($c_2$)

if ( $| a_1 - a_2 | < 0.03490656 $) then we have we have an 
ultrametric-respecting triplet.

\end{itemize}


\begin{thebibliography}{99}

\bibitem{refa1}
J. Austen, {\em Sense and Sensibility} (1811).  Available at: \\
http://www.pemberley.com/etext/SandS

\bibitem{refa2}
J. Austen,   {\em Pride and Prejudice} (1813).  Available at: \\
http://www.pemberley.com/etext/PandP

\bibitem{refa3}
J. Austen,   {\em Persuasion} (1817).  Available at: \\
http://www.pemberley.com/etext/Persuasion

\bibitem{ref2}
J.P. Benz\'ecri,   {\em L'Analyse des Donn\'ees Tome 1, 
La Taxinomie}, 2nd ed., Dunod, Paris, 1979a.

\bibitem{ref3}
J.P. Benz\'ecri,   {\em L'Analyse des Donn\'ees Tome 2, 
Correspondances}, 2nd ed., Dunod, Paris, 1979b.

\bibitem{ref6dom2002}
G.W. Domhoff, ``Using content analysis to study dreams: 
applications and implications for the humanities''. In K. Bulkeley 
(Ed.), {\em Dreams: A Reader on the Religious, Cultural, and 
Psychological Dimensions of Dreaming},  New York: Palgrave,
pp.\ 307--319, 2002.

\bibitem{ref6}
G.W. Domhoff, 
{\em The Scientific Study of Dreams: Neural Networks,
Cognitive Development and Content Analysis}, American Psychological
Association, 2003.

\bibitem{ref6web}
G.W. Domhoff,
``Barb Sanders: our best case study to date, and one that can be built upon'',
http://www2.ucsc.edu/dreams/Findings/barb\_sanders.html
(accessed 1 Jan.\ 2012). 

\bibitem{ref8} 
DreamBank, {\em Repository of Dream Reports}, www.dreambank.net, 2004.


\bibitem{ref9999}
A.K. Hartmann, ``Are ground states of 3D $\pm$ J spin glasses 
ultrametric?'', {\em Europhysics Letters}, 44: 249--254, 1998.

\bibitem{ref10}
I.C. Lerman,  
{\em Classification et Analyse Ordinale des Donn\'ees},
Dunod, Paris, 1981.

\bibitem{ref1414}
I. Matte Blanco, {\em The Unconscious as Infinite Sets: An 
Essay in Bi-Logic}, With a New Foreword by Eric Rayner,
Karnac, London, 1998. (Original version 1975).

\bibitem{ref11} 
F. Murtagh,   ``A survey of recent advances in hierarchical 
clustering algorithms'', {\em The Computer Journal}, 26:
354--359, 1983.

\bibitem{ref13}
F. Murtagh, 
{\em Multidimensional Clustering Algorithms},
Physica-Verlag, W\"urzburg, 1985.

\bibitem{ref14}
F. Murtagh,  ``On ultrametricity, data coding, and computation'',
{\em Journal of Classification}, 21: 167--184, 2004.


\bibitem{ref16} 
F. Murtagh,   {\em 
Correspondence Analysis and Data Coding with Java and R},
Chapman and Hall/CRC Press, New York, 2005b.  

\bibitem{companion}
F. Murtagh,  ``Ultrametric model of mind, I: Review'', 
preprint, 2012. 

\bibitem{ref17}
NTSB, Aviation Accident Database and Synopses, 
National Transport Safety Board,
accessible from http://www.landings.com (2003).


\bibitem{ref18}
J.M. Ockerbloom,  {\em Grimms' Fairy Tales}, 
http://www-2.cs.cmu.edu/$\sim$spok/grimmtmp, 2003.

\bibitem{por}
M.F. Porter,  ``An algorithm for suffix stripping'', 
{\em Program}, 14: 130--137, 1980.

\bibitem{ramaurdou}
R. Rammal, J.C. Angles D'Auriac and B. Doucot, ``On the degree of
ultrametricity'', {\em Le Journal de Physique -- Lettres}, 46: L-945 -- L-952, 
1985.

\bibitem{ref19}
R. Rammal, G.  Toulouse and M.A. Virasoro,  
``Ultrametricity for
physicists'', {\em Reviews of Modern Physics}, 58: 765--788, 1986.


\bibitem{ref20} 
A. Schneider and G.W. Domhoff, {\em The Quantitative Study of Dreams}, 
http://dreamresearch.net, 2004.

\bibitem{refxy}
M. Schweinberger and T.A.B. Snijders,  ``Setting in social networks:
A measurement model'', {\em Sociological Methodology}, 33: 307--342,
2003.

\bibitem{treves}
A. Treves, ``On the perceptual structure of face space'', 
{\em BioSystems}, 40: 189--196, 1997.

\end{thebibliography}
\end{document}